\documentclass[journal]{IEEEtran}
\usepackage{amssymb}
\usepackage{latexsym}
\usepackage{amsmath}
\usepackage{caption}
\usepackage{empheq}

\usepackage{url}
\usepackage{xcolor}
\definecolor{newcolor}{rgb}{.8,.349,.1}

\usepackage{algorithmic}
\begin{document}
\title{A New Point-set Registration Algorithm for Fingerprint Matching}

\author{A.~Pasha~Hosseinbor,
        Renat~Zhdanov,
        and~Alexander~Ushveridze% <-this % stops a space
\thanks{A. P. Hosseinbor and R. Zhdanov are with Bio-Key International Inc., Eagan, MN, USA}

%\thanks{M. Shell was with the Department
%of Electrical and Computer Engineering, Georgia Institute of Technology, Atlanta,

\thanks{A. Ushveridze is with Capella University, Minneapolis, MN, USA.}}% <-this % stops a space
\maketitle

\begin{abstract}
A novel minutia-based fingerprint matching algorithm is proposed that employs iterative global alignment on two minutia sets. The matcher considers all possible minutia pairings and iteratively aligns the two sets until the number of minutia pairs does not exceed the maximum number of allowable one-to-one pairings. The optimal alignment parameters are derived analytically via linear least squares. The first alignment establishes a region of overlap between the two minutia sets, which is then (iteratively) refined by each successive alignment. After each alignment, minutia pairs that exhibit weak correspondence are discarded.  The process is repeated until the number of remaining pairs no longer exceeds the maximum number of allowable one-to-one pairings. The proposed algorithm is tested on both the FVC2000 and FVC2002 databases, and the results indicate that the proposed matcher is both effective and efficient for fingerprint authentication; it is fast and does not utilize any computationally expensive mathematical functions (e.g. trigonometric, exponential). In addition to the proposed matcher, another contribution of the paper is the analytical derivation of the least squares solution for the optimal alignment parameters for two point-sets lacking exact correspondence. 
\end{abstract}


% \fi
\IEEEpeerreviewmaketitle

\section{Introduction}

Fingerprints are the ridge and valley patterns on the tips of human fingers. Due to their uniqueness, fingerprints are widely utilized for personal verification. In fact, fingerprint recognition is one of the most popular biometric technologies in automatic verification systems, and has been extensively employed by forensic experts in criminal investigations.

One of the most important features of a fingerprint are the minutia, which are the points at which the ridges terminate or bifurcate. A detected minutiae in a fingerprint image is characterized by a list of attributes that includes its direction $\theta \in [0,360]$, position $(x,y)$, and type (ridge ending or bifurcation). Fingerprint minutia are widely believed to be the most discriminating and reliable features present in a fingerprint, and for this reason, they are the most widely employed features for fingerprint recognition \cite{maltoni.2009}.

A typical fingerprint recognition algorithm comprises several steps: image acquisition, foreground segmentation, image enhancement and processing, feature extraction, and matching. The last stage, matching, determines whether two different prints belong to the same person or not, and is the main focus of this paper. Many types of fingerprint matchers have been proposed over the years, and they can be divided into three distinct categories: 1) non-minutia-based matching; 2) minutia-based matching; and 3) hybrid matching. Non-minutia methods \cite{jain.2000,xie.2006} compare fingerprints with respect to features extracted from the ridge-furrow pattern (e.g. ridge orientation and frequency, texture). Correlation-based techniques \cite{wilson.2000}, which compare the global pattern of ridges and furrows to see if the ridges in two fingerprint images align, are the most prominent example of non-minutia-based matching. Minutia-based techniques \cite{jiang.2000,jea.2005,wang.2006,sheng.2007,feng.2008}, on the other hand, attempt to align two sets of minutiae points and determine the total number of matched minutia. Hybrid methods \cite{ross.2002,feng.2006,choi.2011} employ both minutia and non-minutia (e.g. ridges) features for matching.

The set of all extracted minutia in a fingerprint image constitutes a point-set, so the problem to be solved by any fingerprint matcher - whether two minutia sets extracted from two different images belong to the same person or not - is tantamount to point pattern matching. Since finger displacement and/or rotation by the user during different image acquisitions frequently arises, an affine transform (i.e. rotation $\theta$, $x$-translation $a$, and $y$-translation $b$) is necessary to register the two minutia sets. In the ideal case, two minutia sets belonging to the same finger would be in exact correspondence, i.e. the two sets are the same size and each minutiae in one set matches to a unique minutiae in the other (one-to-one mapping), so the task then is to determine the (optimal) alignment that minimizes some dissimilarity metric between the two sets, which can be solved analytically via linear least squares \cite{arun.1987,umeyama.1991}. However, in reality, such a situation is rarely encountered in fingerprint biometrics because minutia correspondence is degraded by the following factors:

\begin{enumerate}

\item Finger displacement and rotation may cause part of the fingerprint area to fall outside the sensor's field of view, which results in a smaller overlap between the user's template and input fingerprints. 

\item Both minutia sets may suffer from spurious minutia and be missing genuine minutia, which are caused by poor fingerprint image quality, thereby reducing the overlap between the user's template and input fingerprints.

\item Nonlinear deformations may arise due to the elasticity of the skin, warping the geometry of the ridges. 

\end{enumerate}
Consequently, an exact one-to-one correspondence between two minutia sets rarely exists. Therefore, any alignment scheme for fingerprint images must establish correspondence, not invoke it.

Many different minutia alignment schemes have been proposed, and they can be classified as either local or global. Local minutia alignment-based methods \cite{lee.2002,tico.2003,tong.2005,qi.2005} recover the alignment parameters by choosing a cluster of minutia pairs as a reference  - the pairs that form this reference group are usually taken to be the highest weighted ones - and then aligning the two minutia sets according to this reference. Such an approach properly aligns regions near the reference minutia cluster, but tends to incorrectly align those regions distant from the reference minutia cluster. This is the case because local alignment schemes yield an alignment that is locally strong, but poor in areas far from the reference structure. Global minutia alignment-based methods \cite{zhu.2005,tan.2006,sheng.2007}, on the other hand, seek to evenly align two minutia sets, i.e. finding a transformation that is not biased towards a specific region. 

Irrespective of whether the alignment is global or local, in general, fingerprint matchers execute only a single iteration of minutia alignment, but this may be inadequate to establish correspondence. In many cases, two different prints of the same finger will have little overlap due to noise (e.g. physical condition of the finger, finger pressure upon sensor, and image processing-induced errors). So one application of minutia alignment of a query fingerprint with respect to some template may hardly register the two images. An iterative alignment scheme, which iteratively removes spurious minutia pairings, is desirable since it is more robust to noise.

In this paper, we propose an iterative global alignment-based matcher that considers all possible minutia pairings and iteratively aligns the two minutia sets until the number of pairs does not exceed the maximum number of allowable one-to-one pairings. The optimal alignment parameters are derived analytically via linear least squares. The first alignment establishes a region of overlap between the two point-sets, which is then (iteratively) refined by each successive alignment. After each alignment, minutia pairs that exhibit weak correspondence (i.e. the post-alignment distance of a query minutiae with respect to its potential template minutiae mate exceeds some threshold) are discarded. If a given distance threshold no longer removes any minutia pairs, yet the convergence criterion has not been met, then a more stringent threshold is imposed and the process is repeated until convergence is established (i.e. the number of remaining pairs no longer exceeds the maximum number of allowable one-to-one pairings). A major advantage of the algorithm is its computational efficiency; it is fast and consciously employs as few computationally expensive mathematical functions (e.g. exponential, trigonometric, square root functions) as possible. In addition to the proposed matcher, another contribution of the paper is the analytical derivation of the least squares solution for the optimal alignment parameters for two point-sets lacking exact correspondence. 

The paper is organized as follows. In Section II, we briefly review some related work to emphasize the theoretical contributions of this paper. In Section III, we mathematically formulate our matching algorithm and then describe it numerically. In Section IV, we present and discuss the results of the testing of our algorithm on the FVC2000 and FVC2002 datasets. Lastly, in Section V, we conclude the paper and suggest future directions.

\section{Related Work}
In general, the alignment of two point patterns is a two-part problem; the first problem to be solved is determining the correspondence between the two point-sets, and the second is determining the optimal affine transform that minimizes some dissimilarity metric between the two point-sets. Point pattern matching has been extensively studied in computer vision, and one important class of solutions is linear least squared techniques \cite{arun.1987,umeyama.1991,chang.1997,gold.1998}. Unlike the aforementioned work of \cite{arun.1987,umeyama.1991}, Chang et al. \cite{chang.1997} treated the more general case of two points set of unequal size and do not assume correspondence. They first established correspondence by numerically determining the matching pairs support between the two points sets (i.e. finding an optimal subset of pairings between the two sets), and then derived (analytical) least squared solutions to the transformation parameters that optimally align the optimal subset of pairings. However, their approach is computationally expensive, having quartic polynomial complexity, so rendering it impractical for fingerprint biometrics. Lastly, Gold et al. \cite{gold.1998} determined the optimal affine transform and correspondence simultaneously by numerically solving a constrained least squares problem, but their approach is neither analytical nor computationally efficient. 

Our alignment approach is similar to \cite{arun.1987,umeyama.1991,chang.1997} in that we derive analytical solutions for the optimal affine transform via linear least squares. But unlike these three methods, we do not assume or establish correspondence prior to registration, but use registration to establish correspondence. In order to do this, we must consider all possible pairings between the two minutia sets; otherwise, as will be shown in the next section, the least squares problem becomes ill-posed. To the best of our knowledge, the least square solutions derived in this paper, though remarkably simple, have never been made available in the existing literature. 

\section{Iterative Global Alignment}

In this section, we establish the theoretical and numerical foundations of the proposed algorithm. Our algorithm can be divided into three stages: optimization, alignment, and refinement. 

\subsection{Problem Formulation}
Consider two 2D point-sets ${\bf U}$ and ${\bf V}$ comprising $N_{U}$ and $N_{V}$ singular points, respectively. We can interpret ${\bf U}$ as the query fingerprint image possessing $N_{U}$ minutia, and ${\bf V}$ as the template (reference) fingerprint image possessing $N_{V}$ minutia. Denote the minutia coordinates of each image as

\begin{align*}
{\bf u}_{i}=(x_{i} \; y_{i})^{T} \in {\bf U} \; \; \; i=1,\dots,N_{U} \\
{\bf v}_{k}=(z_{k} \; t_{k})^{T} \in {\bf V} \; \; \; k=1,\dots,N_{V} \\
\end{align*}

We want to register point-set ${\bf U}$ to ${\bf V}$, i.e. align the query image with respect to the template image. There are $N_{U}N_{V}$ possible matching pairs and at most $\min(N_{U},N_{V})$ one-to-one matching pairs. Let $m_{ik}$ denote the weight of a matching pair; the weight can be interpreted as a probability that the points ${\bf u}_{i}$ and ${\bf v}_{k}$ match locally. 

We apply a global rotation and translation to point-set ${\bf U}$:

$$
{\bf u}_{i}^{'} =
\begin{pmatrix}
x_{i}^{'} \\
y_{i}^{'}
\end{pmatrix}
=\begin{pmatrix}
\cos\theta & -\sin\theta \\
\sin\theta & \cos\theta
\end{pmatrix}
\begin{pmatrix}
x_{i} \\
y_{i}
\end{pmatrix}
+\begin{pmatrix}
a \\
b
\end{pmatrix},
$$
where $a$ is the shift along the $x$-direction, $b$ is the shift along the $y$-direction, and $\theta \in [0,2\pi]$ is the rotation.

There are infinitely many possible translations and rotations that could align two given point-sets. The inherent ambiguity of alignment is removed, however, if within the space of alignment parameters we seek to minimize some dissimilarity metric and the minimization is convex; then an optimal alignment is guaranteed. 

The measure of closeness of the transformed point-set ${\bf U'}$ and the template set ${\bf V}$ for a given set of rotation and shift parameters $a$, $b$, and $\theta$ is taken to be the weighted sum of the squared distances between their points:

\begin{equation}
\label{eq:dist}
D({\bf U'},{\bf V};a,b,\theta)=\frac{\sum_{i=1}^{N_{U}}\sum_{k=1}^{N_{V}}m_{ik}({\bf u}_{i}^{'}-{\bf v}_{k})^{T}({\bf u}_{i}^{'}-{\bf v}_{k})}{\sum_{i=1}^{N_{U}}\sum_{k=1}^{N_{V}}m_{ik}}
\end{equation}
We seek the optimal values of parameters $a$, $b$, and $\theta$ that minimize $D({\bf U'},{\bf V};a,b,\theta)$.

\subsection{Initial Minutia Pair Weight}
Before proceeding with optimization, we need to first elaborate on how we compute the pair weights, $m_{ik}$. The importance of the pair weight lies in driving the global registration to the correct alignment. If genuine matching pairs have higher weights than spurious pairs, then global registration has increased likelihood of properly registering the two point-sets; otherwise, misregistration will occur. In order to ensure the pair weights to be as reliable as possible, we incorporate both minutia attributes and local minutia features into the pair weight estimation. The use of local minutia features to aid alignment, such as 1) ridge information associated with a minutiae \cite{jain.1997,he.2003,feng.2008} and 2) features derived from groups of neighboring minutia \cite{vajna.2000,jea.2005,he.2006,chen.2006,feng.2008}, is common in fingerprint recognition.

We define the (initial) probability that two minutia pair up as 

\begin{equation}
\label{eq:weight}
%m_{ik}=e^{-(\delta_{\text{type}_{i},\text{type}_{k}} + (1-q_{i}q_{k}) + S_{\text{NN}_{ik}} )}
m_{ik}=\delta_{\text{type}_{i},\text{type}_{k}}*q_{i}q_{k}*S_{\text{NN}_{ik}}
\end{equation} 
Here, $\text{type}_{i}$ and $\text{type}_{k}$ indicate the minutia type (ending or bifurcation) of the $i^{th}$ and $k^{th}$ minutiae in the query and template images, respectively, and

\begin{align}
%\[
  \delta_{\text{type}_{i},\text{type}_{k}} = \left\{\def\arraystretch{1.2}%
  \begin{array}{@{}c@{\quad}l@{}}
   0.5 &  \text{type}_{i} \neq \text{type}_{k}\\
    1 & \text{type}_{i} = \text{type}_{k}\\
  \end{array}\right.
%\]
\end{align}
Such a formulation for the pair weightings incorporates the effects of pairs formed by differing minutia types on the metric minimization; assuming that endings may only map to endings and bifurcations to bifurcations is not prudent because image processing can convert a genuine minutiae ending into a bifurcation and vice versa (smoothing a ridge in a direction not parallel to the ridge orientation can cause this). 
%Note that Eq. (\ref{eq:weight}) does not distinguish between pairs comprising differing minutia types and those formed by the same type, because minutia type can be easily interchanged by noise and image processing.

The terms $q_{i}$ and $q_{k}$ are the minutia quality scores ($q \in [0,1]$) of the $i^{th}$ query minutiae and $k^{th}$ template minutiae, respectively. The quality score is a measure of the certitude that a given minutiae is genuine.  

The term $S_{\text{NN}_{ik}}$ is related to the nearest neighbor ("NN") data of the $i^{th}$ query minutiae and $k^{th}$ template minutiae. The nearest neighbor refers to the minutiae closest to the minutiae of interest along some angular direction (e.g. the minutiae closest to the reference minutiae along a $45^{\circ}$ ray oriented with respect to the reference minutiae that joins the two), and its data encompasses both the Euclidean distance and minutia angle difference between the two minutia. We acquire the nearest neighbor data in each of the eight angular octants (i.e. 0, 45, 90, 135, 180, 225, 270, 315$^{\circ}$) about the reference minutiae \cite{incits.2004}. The purpose of the nearest neighbor data is to assess whether a potential minutia pair exhibit radial and angular invariance; if query minutiae $i$ corresponds to template minutiae $k$, then the radial distance and minutia angle difference between $i$ and its nearest neighbor should be the same as those between $k$ and its near neighbor. Thus, it allows Eq. (\ref{eq:weight}) to weigh more heavily those minutia pairs that exhibit both radial and angular invariance with respect to their nearest neighbors.
 
$S_{\text{NN}_{ik}}$ is computed in the following way:

\begin{enumerate}

\item A corresponding octant between query minutiae $i$ and template minutiae $k$ arises when the $l^{th}$ octant ($l=1,2,...,8$) of both minutia $i$ and $k$ contains a nearest neighbor. Count the number of corresponding octants, which we denote as $n_{\text{octants}}$. If a nearest neighboring minutiae exists in octant $l$, compute both the Euclidean distance and (minutia) angle difference between it and the reference minutiae. The distance and angle difference are denoted as $d_{il}$ and $\psi_{il}$, respectively, for minutiae $i$, and $d_{kl}$ and $\psi_{kl}$ for minutiae $k$.

\item Query minutiae $i$ and template minutiae $k$ have a matching $l^{th}$ octant if and only if $ | d_{il} - d_{kl} | \leq t_{d} $ and $| \psi_{il} - \psi_{kl} | \leq t_{\psi}$, where $t_{d}$ and $t_{\psi}$ are empirical thresholds that ideally should be close to 0. Denote the number of matching octants between minutia pair $(i,k)$ as $n_{\text{matching},ik}$. 

\end{enumerate}
Thus, we define $S_{\text{NN}_{ik}}$ as

%\small
%\begin{align}
%S_{\text{NN}_{ik}} = \left\{\def\arraystretch{1.2}
%\begin{array}{@{}c@{\quad}l@{}}
%   1 - \frac{n_{\text{matching},ik}}{n_{\text{octants}}} & n_{\text{octants}} \neq 0\\
%    0.5 & n_{\text{octants}} = 0\\
%  \end{array}\right.
%\end{align}
%\normalsize 
%In other words, $S_{\text{NN}_{ik}}$ is the fraction of corresponding octants between query minutiae $i$ and template minutiae $k$ that are not matching; the smaller $S_{\text{NN}_{ik}}$ is, the greater the weight, $m_{ik}$. 
\small
\begin{align}
S_{\text{NN}_{ik}} = \left\{\def\arraystretch{1.2}
\begin{array}{@{}c@{\quad}l@{}}
    \frac{n_{\text{matching},ik}}{n_{\text{octants}}} & n_{\text{octants}} \neq 0\\
    0.5 & n_{\text{octants}} = 0\\
  \end{array}\right.
\end{align}
\normalsize 
In other words, $S_{\text{NN}_{ik}}$ is the fraction of corresponding octants between query minutiae $i$ and template minutiae $k$ that are matching; the smaller $S_{\text{NN}_{ik}}$ is, the smaller the weight, $m_{ik}$.

%Hence, a pair formed by two minutia of the same type and whose product of their quality scores is near 1 will have a higher weight than a pair formed by minutia having either differing types or low quality scores. 

\subsection{Numerical Implementation}
{\bf Step 1: Least Squares Minimization}

Recall that we seek the optimal values of parameters $a$, $b$, and $\theta$ that minimize $D({\bf U'},{\bf V};a,b,\theta)$:

\begin{equation}
\label{eq:minimize}
(\hat{a} \; \hat{b} \; \hat{\theta})=\min_{a,b,\theta}D({\bf U'},{\bf V};a,b,\theta)
\end{equation}
Before proceeding, we define the weighted averages of the minutia coordinates as

\begin{align*}
\overline{x}=\frac{\sum_{i=1}^{N_{U}}\sum_{k=1}^{N_{V}}m_{ik}x_{i}}{\sum_{i=1}^{N_{U}}\sum_{k=1}^{N_{V}}m_{ik}} \\
\overline{y}=\frac{\sum_{i=1}^{N_{U}}\sum_{k=1}^{N_{V}}m_{ik}y_{i}}{\sum_{i=1}^{N_{U}}\sum_{k=1}^{N_{V}}m_{ik}} \\
\overline{z}=\frac{\sum_{i=1}^{N_{U}}\sum_{k=1}^{N_{V}}m_{ik}z_{k}}{\sum_{i=1}^{N_{U}}\sum_{k=1}^{N_{V}}m_{ik}} \\
\overline{t}=\frac{\sum_{i=1}^{N_{U}}\sum_{k=1}^{N_{V}}m_{ik}t_{ik}}{\sum_{i=1}^{N_{U}}\sum_{k=1}^{N_{V}}m_{ik}} 
\end{align*}
Differentiating Eq. (\ref{eq:dist}) with respect to alignment parameters $a$, $b$, and $\theta$ and then minimizing yields the optimal alignment parameters:

%\begin{equation*}
%\begin{aligned}
%%\boxed{
%\hat{\theta}= \text{atan$2$} \left(-w_{4},w_{1} \right) \\
%\hat{a}=\overline{z}+\overline{y}\sin\hat{\theta}-\overline{x}\cos\hat{\theta} \\ 
%\hat{b}=\overline{t}-\overline{y}\cos\hat{\theta}-\overline{x}\sin\hat{\theta}
%%}
%\end{aligned}
%\end{equation*}

%\begin{subequations}
\begin{empheq}[box=\fbox]{align}
\label{eq:opt_sol}
\hat{\theta}&= \text{atan$2$} \left(-w_{4},w_{1} \right) \nonumber \\ 
\hat{a}&=\overline{z}+\overline{y}\sin\hat{\theta}-\overline{x}\cos\hat{\theta} \\ 
\hat{b}&=\overline{t}-\overline{y}\cos\hat{\theta}-\overline{x}\sin\hat{\theta} \nonumber
\end{empheq}
%\end{subequations}

where

\begin{equation*}
\begin{aligned}
w_{1}=\sum_{i=1}^{N_{U}}\sum_{k=1}^{N_{V}}m_{ik}\left[(z_{k}-\overline{z})x_{i}+(t_{k}-\overline{t})y_{i}\right] \\
w_{4}=\sum_{i=1}^{N_{U}}\sum_{k=1}^{N_{V}}m_{ik}\left[(z_{k}-\overline{z})y_{i}-(t_{k}-\overline{t})x_{i}\right] 
\end{aligned}
\end{equation*}
The full derivation of Eq. (\ref{eq:opt_sol}) is presented in the Appendix. Eq. (\ref{eq:opt_sol}) is only relevant to the case of inexact correspondence between two point-sets, but it is identical in form to the least squares solution for exact correspondence \cite{umeyama.1991}. Although remarkably simple, we were unable to find any similar derivation of Eq. (\ref{eq:opt_sol}) in the available literature. 

The double summation in Eq. (\ref{eq:dist}) is equivalent to summing across all possible pairings between point-sets ${\bf U}$ and ${\bf V}$. For this reason, it will be useful to introduce a queue of all possible minutia pairings between point-sets $U$ and $V$, which we denote as ${\bf M}$. It will keep track of which minutia are discarded or kept after each alignment. Initially, ${\bf M}$ contains $N_{U}N_{V}$ elements.

{\bf Step 2: Alignment} \hfill \break
The optimal alignment parameters represent the rotation and shift minimizing the (weighted) averaged squared distance between two points sets. After alignment, the squared (radial) distance between two minutia forming a potential pair is 

\scriptsize
\begin{equation}
% D_{ik}=\sqrt{(x_{i}\cos\hat{\theta}-y_{i}\sin\hat{\theta} + \hat{a} - z_{k})^{2} + (x_{i}\sin\hat{\theta} + y_{i}\cos\hat{\theta} + \hat{b} - t_{k})^{2}}
D_{ik}^{2}=(x_{i}\cos\hat{\theta}-y_{i}\sin\hat{\theta} + \hat{a} - z_{k})^{2} + (x_{i}\sin\hat{\theta} + y_{i}\cos\hat{\theta} + \hat{b} - t_{k})^{2}
\end{equation}
\normalsize 
Likewise, we can compute the minutia angle difference between the two minutia. Let $\alpha_{i}$ and $\alpha_{k}$ denote the minutia angles ($\alpha \in [0,2\pi]$) of the $i^{th}$ and $k^{th}$ minutiae in the query and template images, respectively. Then the post-alignment angular difference is 

\begin{equation}
\Theta_{ik}=\min(|\alpha_{i}+\hat{\theta}-\alpha_{k}|, 360-|\alpha_{i}+\hat{\theta}-\alpha_{k}|)
\end{equation}
Hence, the total post-alignment displacement of the $i^{th}$ query minutiae with respect to the $k^{th}$ template minutiae is 

\begin{equation}
% \Delta_{ik} = c_{1}D_{ik} + c_{2}\Theta_{ik},
\Delta_{ik} = c_{1}D_{ik}^{2} + c_{2}\Theta_{ik}^{2},
\end{equation}
where $c_{1}$ and $c_{2}$ are normalization terms that address the difference in units between radial and angular displacement. If a pair constitutes a genuine match, then ideally $\Delta_{ik}$ will be small, and we will want to keep it. And if the pair is spurious, then $\Delta_{ik}$ will be large, and we will want to discard it. To do that, we need to compare each pair's $\Delta_{ik}$ to some threshold. 

Let ${\bf T}$ the $n \times 1$ vector of thresholds, where $T_{1} > T_{2} > \cdots > T_{n}$. For simplicity, we take the elements of ${\bf T}$ to be evenly spaced, so $T_{j}=T_{1}-(j-1)c$, where $j=1,...,n$ and $c$ is some real constant. We start out with a large threshold, $T_{1}$, so to remove the "most obviously bad pairs", i.e. those that have very large $\Delta_{ik}$. If for a given pair

\begin{equation}
\Delta_{ik} > T_{1},
\end{equation}
then the pair is an outlier and we remove from it from the queue ${\bf M}$. Otherwise, we recompute its weight as 

\begin{equation}
% m_{ik}=e^{-\sqrt{(c_{1}D_{ik})^{2} + (c_{2}\Theta_{ik})^{2}}}
%m_{ik}=e^{-\sqrt{\Delta_{ik}}}
m_{ik}=1-\frac{\Delta_{ik}}{T_{1}}
\end{equation}

{\bf Step 3: Refinement} \hfill \break
Upon iterating across every pair in ${\bf M}$, we count the number of pairs left in the queue. If no pairs have been removed, this is indicative of the threshold being too large, so we repeat Step 2, but using the next largest threshold in ${\bf T}$ (e.g. If $T_{1}$ failed to remove any pairs, then use $T_{2}$, and etc). 

If pairs have been removed, then we need to check if the convergence criterion has been met. We define convergence as when the number of pairs does not exceed the maximum number of allowable one-to-one matchings, i.e. $\text{length}({\bf M}) < \min{\left( N_{U}, N_{V} \right)}$. If this happens to be the case, then we are done. Otherwise, we repeat Steps 1 $\&$ 2 using the newly shortened queue ${\bf M}$ and the new weights. 

\subsection{Numerical Implementation}
We now summarize our algorithm as a pseudocode:

\begin{algorithmic}[1]
\FOR {each threshold $T_{n}$} \STATE{
\WHILE {number of pairs $> \min(N_{U},N_{V})$} \STATE{Align query fingerprint to template
\FOR {each minutia pair $M_{j}$ in queue {\bf \textit{M}}} \STATE{compute weighted sum, $\Delta_{j}$, of radial and angular displacements
\IF {$\Delta_{j} > T_{n}$} \STATE{ remove minutia pair $M_{j}$ from queue {\bf \textit{M}}} 
\ELSE \STATE{compute new weight of $M_{j}$} 
\ENDIF 
}
\ENDFOR
\IF {no minutia pairs are removed} \STATE{break} \COMMENT{Current threshold inadequately non-discriminative; move to next threshold}
\ENDIF
%\IF {number of minutia pairs $ \leq > \min\{N_{U},N_{V}\}$} \STATE{break} \COMMENT{Solution has converged} \ENDIF
}
\ENDWHILE
}
\ENDFOR
\end{algorithmic}

\subsection{Uncoupled Weights: An Ill-Posed Problem }
We now examine a specific case that makes the minimization problem described by Eq. (\ref{eq:minimize}) ill-posed, i.e. no unique solution exists. Lets assume that the event of query minutiae $i$ paring up with template minutiae $k$ is independent of template minutiae $k$ pairing up with query minutiae $i$, which is tantamount to the weight term, $m_{ik}$, becoming separable, i.e. $m_{ik}=\sigma_{i}\gamma_{k}$. In this case, we have

\small
\begin{align*}
w_{1}=\sum_{i=1}^{N_{U}}\sigma_{i}x_{i}\sum_{k=1}^{N_{V}}\gamma_{k}(z_{k}-\overline{z}) + \sum_{i=1}^{N_{U}}\sigma_{i}y_{i}\sum_{k=1}^{N_{V}}\gamma_{k}(t_{k}-\overline{t})y_{i}=0 \\
w_{4}=\sum_{i=1}^{N_{U}}\sigma_{i}y_{i}\sum_{k=1}^{N_{V}}\gamma_{k}(z_{k}-\overline{z}) - \sum_{i=1}^{N_{U}}\sigma_{i}x_{i}\sum_{k=1}^{N_{V}}\gamma_{k}(t_{k}-\overline{t})=0
\end{align*}
\normalsize

%\begin{eqnarray*}
%w_{1}=\sum_{i=1}^{N_{U}}\sum_{k=1}^{N_{V}}\sigma_{i}\gamma_{k}\left[(z_{k}-\overline{z})x_{i}+(t_{k}-\overline{t})y_{i}\right]=0 \\
%w_{4}=\sum_{i=1}^{N_{U}}\sum_{k=1}^{N_{V}}\sigma_{i}\gamma_{k}\left[(z_{k}-\overline{z})y_{i}-(t_{k}-\overline{t})x_{i}\right]=0
%\end{eqnarray*}
which follows from the fact that sum of the deviations from the mean is always $0$. As a result, no unique solution exists for the optimization problem. A special sub-case is when the weight term is fixed for all possible pairs, i.e. $m_{ik}=c$ for all $i$ and $k$, where $c$ is a real constant. Thus, a unique solution for the minimization problem given by Eq. (\ref{eq:minimize}) is guaranteed if and only if the weight term is coupled between the two point-sets. 

Coupled weights, $m_{ik}$, make the minimization problem well-posed because they establish a quasi-correspondence between the two point-sets; higher weighted pairs are more likely to exhibit correspondence. The probability that two minutia pair up is a function of each minutiae's attributes (e.g. ridge and nearest neighbor information), which are not independent of another; if the pair is genuine, then their attributes will be similar, while if the pair is erroneous, then their attributes will be different. If the weights are not coupled, - that is, the attributes of each of the two minutia forming a pair is independent of one another - then no correspondence exists between the two point-sets, so nothing is known about how the minutia pair up. Thus, there will be infinitely many possible solutions for the alignment.

\section{Experiments and Results}

We tested our algorithm on the FVC2000 and FVC2002 databases, which each contain three real (i.e. non-synthetic) datasets comprising 800 gray-level fingerprint images. The 800 fingerprint images in each dataset are acquired from 100 different subjects, eight times each. We perform all one-to-one comparisons for each dataset, which means 2800 genuine and 316800 imposter comparisons are carried out for each dataset (hence, 319600 total comparisons are performed for a given dataset). 

Algorithmic parameters are $t_{d}=10$; $t_{\psi}=20^{\circ}$; and $T_{1}=24$, which is decremented by a step-size of four. Prior to the matching stage, all images are binarized and thinned using in-house algorithms. The results reported in this section were obtained on a PC with an Intel(R) Core(TM) i7-3930K processor at 3.2 GHz.

The equal error rates (EER) and average comparison time for each dataset are displayed in Table \ref{table:Results}. On the good quality datasets (FVC2000\_1, FVC2000\_2, FVC2002\_1, and FVC2002\_2), the proposed matcher consistently archives an EER of less one percent. However, it does not perform as well on the two noisy datasets (FVC2000\_3 and FVC2002\_3), which implies that the initial weights we are using are not robust to noise. Regarding the FVC2002 database, the proposed matcher performs as well as or better (in terms of EER) than the matchers proposed in \cite{feng.2006,sheng.2007,choi.2011} on each dataset. The matcher proposed in \cite{xie.2006} achieves better accuracy for FVC2002, but their EER calculation is based on approximately one-tenth of the total number of imposter comparisons. 

Note that the only computationally intricate mathematical function utilized by the matcher is the $\arctan2$ function, which can be replaced by a lookup table. This fact, in addition to its fast compare time, suggests that the proposed matcher may be compatible (after some modifications) for embedded biometric systems. 

\section{Conclusion}

A novel minutia-based matcher has been proposed in this paper. It considers all possible minutia pairings between two minutia sets, and unlike other matchers, it iteratively aligns the two sets until the number of minutia pairs does not exceed the maximum number of allowable one-to-one pairings. The optimal alignment parameters are derived analytically via linear least squares. The first alignment establishes a region of overlap between the two point-sets, which is then (iteratively) refined by each successive alignment. After each alignment, minutia pairs that exhibit weak correspondence are discarded.  The process is repeated until the number of remaining pairs no longer exceeds the maximum number of allowable one-to-one pairings. Experimental results on the FVC2000 and FVC2002 databases show that the proposed matcher is both effective and efficient for fingerprint authentication. In addition to the proposed matcher, another contribution of the paper is the analytical derivation of the least squares solution for the optimal alignment parameters for two point-sets lacking exact correspondence. 

The current algorithm can be improved, in terms of accuracy, by formulating more robust and discriminative weights. Since the proposed matcher utilizes no computationally intricate mathematical functions and is fast, another avenue of future research is employing it in an embeddable biometric environment.

\begin{table}
%\begin{minipage}[b]{0.5\linewidth}
\caption{Results on FVC2000 and FVC2002 Databases}
\centering
\begin{tabular}{| c | c | c | c|}
\hline
%\small
Dataset & Image Size & Comparison Time (ms) & EER ($\%$) \\ \hline
FVC2000\_1 & $300 \times 300$ & 2.63 & 0.818 \\ \hline
FVC2000\_2 & $256 \times 364$ & 3.45  & 0.654  \\ \hline
FVC2000\_3 & $448 \times 478$ & 6.68 & 4.55  \\ \hline
%FVC2000\_4 & 24 & 45.6 & 15.2  \\ \hline
FVC2002\_1 & $388 \times 374$ & 2.89 & 0.890 \\ \hline
FVC2002\_2 & $296 \times 560$ & 2.63 & 0.462 \\ \hline
FVC2002\_3 & $300 \times 300$ & 2.98 & 3.51 \\ \hline
%FVC2002\_4 &  &  &  \\ \hline
\end{tabular}
%\end{minipage}
\label{table:Results}
\end{table}
\normalsize

\appendix[Derivation of Linear Least Squares Solution]

Optimizing Eq. (\ref{eq:dist}) with respect to $a$ and $b$, i.e. differentiating and setting to zero, gives

\begin{equation}
\label{eq:ab}
\begin{aligned}
\hat{a}=\overline{z}+\overline{y}\sin\theta-\overline{x}\cos\theta \\ 
\hat{b}=\overline{t}-\overline{y}\cos\theta-\overline{x}\sin\theta
\end{aligned}
\end{equation}
Optimizing Eq. (\ref{eq:dist}) with respect to $\theta$ gives

\begin{align}
\label{eq:dtheta}
\sin\theta\sum_{i=1}^{N_{U}}\sum_{k=1}^{N_{V}}m_{ik} \left[(z_{k}-a)x_{i}+(t_{k}-b)y_{i}\right] + \\ \nonumber \cos\theta\sum_{i=1}^{N_{U}}\sum_{k=1}^{N_{V}}m_{ik} \left[(z_{k}-a)y_{i}-(t_{k}-b)x_{i}\right] =0
\end{align}
\normalsize
Substituting Eqs. (\ref{eq:ab}) for $a$ and $b$ into Eq. (\ref{eq:dtheta}) yields

\begin{equation*}
w_{1}\sin\theta+w_{4}\cos\theta=0,
\end{equation*}
whose solution is

\begin{equation*}
\hat{\theta}=\text{atan$2$} \left(-w_{4},w_{1} \right)
\end{equation*}
Thus, the optimal alignment parameters are

\begin{equation*}
\begin{aligned}
%\boxed{
\hat{\theta}=\text{atan$2$} \left(-w_{4},w_{1} \right) \\
\hat{a}=\overline{z}+\overline{y}\sin\hat{\theta}-\overline{x}\cos\hat{\theta} \\ 
\hat{b}=\overline{t}-\overline{y}\cos\hat{\theta}-\overline{x}\sin\hat{\theta}
%}
\end{aligned}
\end{equation*}

In the case of exact correspondence, the weights in Eq. (\ref{eq:dist}) become $m_{ik}=w_{i}\gamma_{ik}$, where

\begin{align*}
%\[
  \gamma_{ik} = \left\{\def\arraystretch{1.2}%
  \begin{array}{@{}c@{\quad}l@{}}
   0 &  \text{minutia $i$ and $k$ do not correspond} \\
    1 & \text{minutia $i$ and $k$ do correspond} \\
  \end{array}\right.
%\]
\end{align*}

% or


%\appendices

%\section{}

%\section*{Acknowledgment}

\ifCLASSOPTIONcaptionsoff
  \newpage
\fi



\bibliographystyle{IEEEtran}
\bibliography{refs}


%\end{IEEEbiography}


%\newpage



%\vfill


\end{document}